\documentclass[11pt,a4paper]{article}
\usepackage[hyperref]{acl2019}
\usepackage{graphicx}
\usepackage{times}
\usepackage{latexsym}
\usepackage{amssymb}
\usepackage{amsmath}
\usepackage{multirow}
\usepackage{color}

\def\vector#1{\mbox{\boldmath $#1$}}
\def\fig#1{\figurename~\ref{fig:#1}}
\def\tbl#1{\tablename~\ref{tab:#1}}
\def\Sec#1{Section~\ref{sec:#1}}

\aclfinalcopy
\def\aclpaperid{26}

\usepackage{breakurl}
\hypersetup{breaklinks=true}

\title{A Simple but Effective Method to Incorporate Multi-turn Context \\with BERT for Conversational Machine Comprehension}

\author{
    Yasuhito Ohsugi \\\And
    Itsumi Saito\\\And
    Kyosuke Nishida \\
    NTT Media Intelligence Laboratories, NTT Corporation \\ 
    \texttt{yasuhito.ohsugi.va@hco.ntt.co.jp} \\\And
    Hisako Asano \\\And
    Junji Tomita \\
}

\date{}

\begin{document}
\maketitle

\begin{abstract}
Conversational machine comprehension (CMC) requires understanding the context of multi-turn dialogue.
Using BERT, a pre-training language model, has been successful for single-turn machine comprehension, while modeling multiple turns of question answering with BERT has not been established because BERT has a limit on the number and the length of input sequences.
In this paper, we propose a simple but effective method with BERT for CMC.
Our method uses BERT to encode a paragraph independently conditioned with each question and each answer in a multi-turn context.
Then, the method predicts an answer on the basis of the paragraph representations encoded with BERT.
The experiments with representative CMC datasets, QuAC and CoQA, show that our method outperformed recently published methods (+0.8 F1 on QuAC and +2.1 F1 on CoQA).
In addition, we conducted a detailed analysis of the effects of the number and types of dialogue history on the accuracy of CMC, and we found that the gold answer history, which may not be given in an actual conversation, contributed to the model performance most on both datasets.
\end{abstract}

\section{Introduction}
Single-turn machine comprehension (MC) has been studied
as a question answering method \citep{BiDAF, DrQA, QANet, Lewis2018generative}.
Conversational artificial intelligence (AI) such as Siri and Google Assistant requires
answering not only a single-turn question but also multi-turn questions in a dialogue. 
Recently, two datasets, QuAC \citep{QuAC} and CoQA \citep{CoQA}, were released
to answer sequential questions in a dialogue by comprehending a paragraph.
This task is called conversational machine comprehension (CMC) \citep{FlowQA},
which requires understanding the context of multi-turn dialogue that consists of the question and answer history.

Learning machine comprehension models requires a lot of question answering data.
Therefore, transfer learning from pre-training language models based on a large-scale unlabeled corpus is useful for improving the model accuracy.
In particular, BERT \citep{BERT} achieved state-of-the-art results when performing various tasks including the single-turn machine comprehension dataset SQuAD \citep{SQuAD_v1}.
BERT takes a concatenation of two sequences as input during pre-training and can capture the relationship between the two sequences.
When adapting BERT for MC, we use a question and a passage as input and fine-tune the pre-trained BERT model to extract an answer from the paragraph.
However, BERT can accept only two sequences of 512 tokens and thus cannot handle CMC naively.

\citet{SDNet} proposed a method for CMC that is based on an architecture for single-turn MC and uses BERT as a feature-based approach. To convert CMC into a single-turn MC task, the method uses a reformulated question, which is the concatenation of the question and answer sequences in a multi-turn context with a special token. It then uses BERT to obtain contextualized embeddings for the reformulated question and paragraph, respectively.
However, it 
cannot use BERT to capture the interaction between each sequence in the multi-turn context and the paragraph.


In this paper, we propose a simple but effective method for CMC based on a fine-tuning approach with BERT.
Our method consists of two main steps. 
The first step is contextual encoding where BERT is used for 
independently obtaining paragraph representations conditioned with the current question, each of the previous questions, and each of the previous answers. 
The second step is answer span extraction,
where the start and end position of the current answer are predicted
based on the concatenation of the paragraph representations encoded in the previous step.

The contributions of this paper are as follows:
\begin{itemize}
    \item
    We propose a novel method for CMC based on fine-tuning BERT by regarding the sequences of the questions and the answers as independent inputs.
    \item
    The experimental results show that our method outperformed published methods on both QuAC and CoQA.
    \item
   We found that the gold answer history contributed to the model performance most by analyzing the effects of dialogue history.
\end{itemize}

\section{Task Definition} \label{sec:task_definition}
In this paper, we define the CMC task as follows:
\begin{itemize}
\item \textbf{Input}: Current question $Q_{i}$, paragraph $P$,
previous questions $\{Q_{i-1}, ... , Q_{i-k}\}$,
and previous answers $\{A_{i-1}, ..., A_{i-k}\}$
\item \textbf{Output}: Current answer $A_{i}$ and type $T_{i}$
\end{itemize}
where $i$ and $k$ denote the turn index in the dialogue and the number of considered histories (turns), respectively.
Answer $A_{i}$ is a span of paragraph $P$.
Type $T_{i}$ is \textit{SPAN}, \textit{YES}, \textit{NO}, or \textit{UNANSWERABLE}.

\section{Pre-trained Model} \label{sec:pre_trained_model}
BERT is a powerful language representation model \citep{BERT},
which is based on bi-directional Transformer encoder \citep{Transformer}.
BERT can obtain language representation by unsupervised pre-training with a huge data corpus
and by supervised fine-tuning, and it can achieve outstanding results in various NLP tasks 
such as sentence pair classification, single sentence tagging, and single-turn machine comprehension.

Here, we explain how to adapt BERT for single-turn machine comprehension tasks such as SQuAD \citep{SQuAD_v1}.
In SQuAD, a question and a paragraph containing the answer are given,
and the task is to predict the answer text span in the paragraph.
In the case of using BERT for SQuAD,
after the special classification token [CLS] is added in front of the question,
the question and the paragraph are concatenated with special tokens [SEP] into one sequence.
The sequence is inputted to BERT with segment embeddings and positional embeddings.
Then, the final hidden state of BERT is converted to the probabilities of answer span 
by a linear layer and softmax function.
The fined-tuned BERT for the SQuAD dataset can capture the relationship between one question and one paragraph
so that BERT achieved state-of-the-art performance on the SQuAD.
However, BERT itself cannot be used for a task requiring multiple queries or multiple paragraphs,
because BERT can accept only two segments in one input sequence.
This limitation can be a problem for the CMC task
because there are multi-turn questions about the same paragraph.

\section{Proposed Method} \label{sec:proposed_method}
\begin{figure}
    \centering
    \includegraphics[width=8cm]{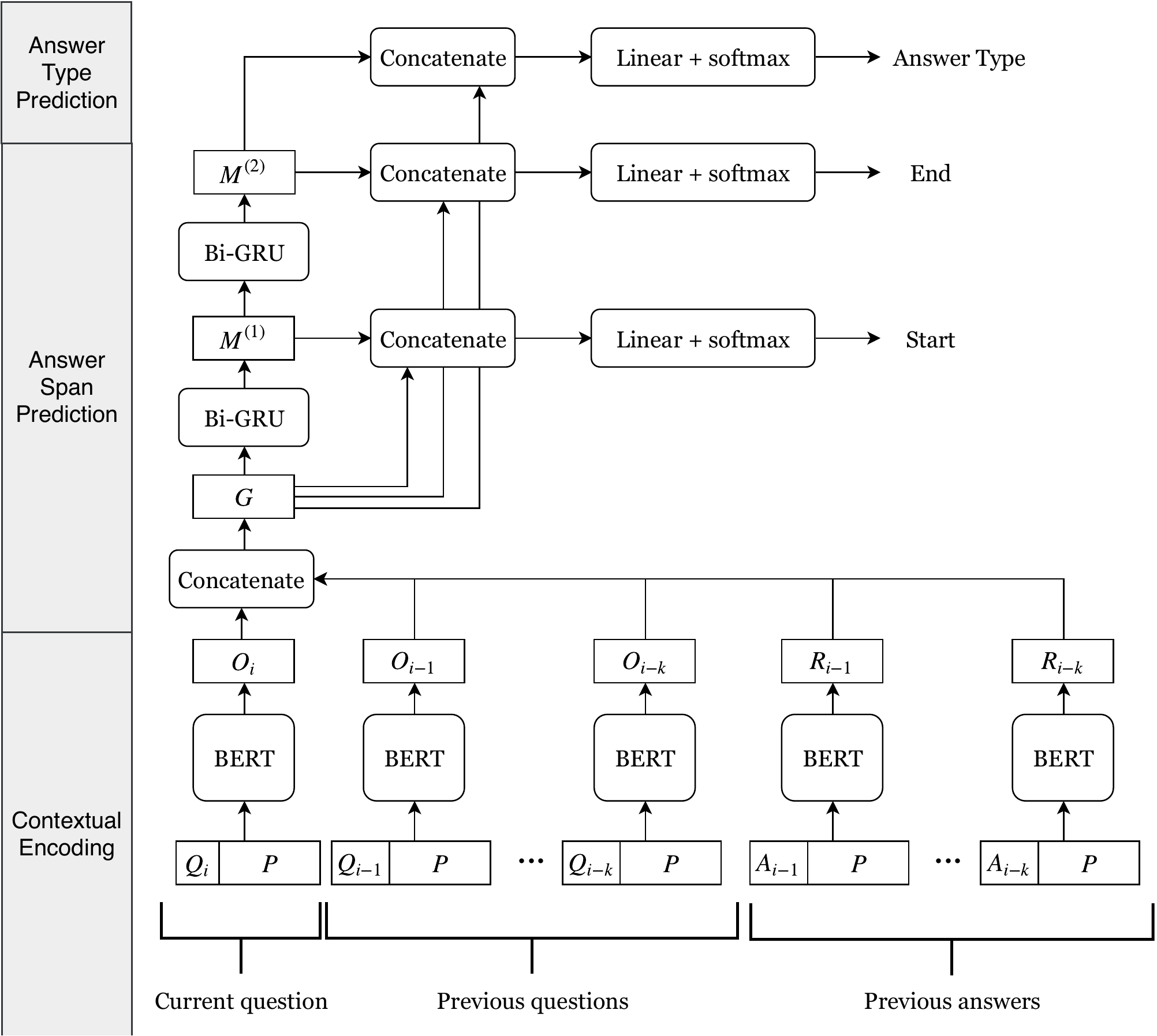}
    \caption{Our model}
    \label{fig:network_architecture}
\end{figure}
In the CMC task, it is necessary to consider
not only the current question $Q_{i}$ but also the question history $\{Q_{i-1}, ... , Q_{i-k}\}$
and the answer history $\{A_{i-1}, ..., A_{i-k}\}$.
We propose a method of modeling the current question, question history, and answer history
by using BERT (\fig{network_architecture}).
Our method consists of two steps: contextual encoding and answer span prediction.
On top of that, answer type is predicted only in the case of CoQA (see \Sec{answer_type_prediction}).

\subsection{Contextual Encoding}
In this step, we use BERT to encode not only the relationship between the current question and the paragraph
but also the relationship between the history and the paragraph.
We define the method of extracting features by using BERT as follows,
\begin{equation}
    \vector{z} = f(\textrm{BERT}(x, y|\theta)),
\end{equation}
where $x$, $y$, and $\vector{z}$ denote the input query sequence, input paragraph sequence, and output feature, respectively.
The function $\textrm{BERT}(\cdot)$ outputs BERT's $d$-dimensional final hidden states with parameters $\theta$,
and the function $f(\cdot)$ extracts features corresponding to the segment of the paragraph in the final hidden states.
Namely, if input paragraph text $y$ has $T$ tokens, then, $\vector{z} \in \mathbb{R}^{d \times T}$.
This step consists of three parts, and each part shares the BERT parameters $\theta$.
First, we encode the current question as follows,
\begin{equation}
    \vector{O_{i}} = f(\textrm{BERT}(Q_{i}, P|\theta)).
\end{equation}
Second, we encode the question history $\{Q_{i-1}, ... , Q_{i-k}\}$ in the same manner.
\begin{equation}
    \vector{O_{i-l}} = f(\textrm{BERT}(Q_{i-l}, P|\theta)),
\end{equation}
where $l$ denotes the index of the previous context.
Last, we encode the answer history $\{A_{i-1}, ..., A_{i-k}\}$.
Note that previous answer $A_{i-l}$ is given as text,
even if the current answer is predicted as the span of the paragraph.
The encoded feature can be obtained as follows,
\begin{equation}
    \vector{R_{i-l}} = f(\textrm{BERT}(A_{i-l}, P|\theta)).
\end{equation}

\subsection{Answer Span Prediction}
In this step, the current answer span is predicted.
Let $s_{i}$ and $e_{i}$ represent the start index and the end index, respectively.
First, the output features of the previous step are concatenated as follows,
\begin{equation}
    \vector{G} = [\vector{O_{i}}; \vector{O_{i-1}};...; \vector{O_{i-k}}; \vector{R_{i-1}}; ...; \vector{R_{i-k}}],
\end{equation}
where $[;]$ is vector concatenation across row and $\vector{G} \in \mathbb{R}^{(2k+1)d \times T}$.
Then, $\vector{G}$ is passed to BiGRU over tokens and converted to $\vector{M^{(1)}} \in \mathbb{R}^{2d \times T}$.
To predict the start index $s_{i}$, the probability distribution is calculated by,
\begin{equation}
    p^{\textrm{s}} = \textrm{softmax}\left( \vector{w}_{1}^{\top} [\vector{G}; \vector{M}^{(1)}] + \vector{b}_{1}\right),
\end{equation}
where $\vector{w}_{1}$ and $\vector{b}_{1} \in \mathbb{R}^{(2k+3)d}$ are trainable vectors.
Next, to predict the end index $e_{i}$,
$\vector{M^{(1)}}$ is passed to another BiGRU over tokens and converted to $\vector{M^{(2)}} \in \mathbb{R}^{2d \times T}$.
Then, the probability distribution is calculated by
\begin{equation}
    p^{\textrm{e}} = \textrm{softmax}\left( \vector{w}_{2}^{\top} [\vector{G}; \vector{M}^{(2)}] + \vector{b}_{2}\right),
\end{equation}
where $\vector{w}_{2}$ and $\vector{b}_{2} \in \mathbb{R}^{(2k+3)d}$ are trainable vectors.

\subsection{Answer Type Prediction} \label{sec:answer_type_prediction}
Some questions should be simply answered as "yes" or "no" and not answered as a rationale text.
To address these questions, the probability of the answer type is calculated as follows,
\begin{equation}
    p^{\textrm{ans}} = \left[
        \textrm{softmax}\left( \vector{w}_{3}^{\top} [\vector{G}; \vector{M}^{(2)}] + \vector{b}_{3}\right)
        \right]_{e_{i}},
\end{equation}
where $\vector{w}_{3}$ and $\vector{b}_{3} \in \mathbb{R}^{(2k+3)d}$ are trainable vectors
and $e_{i}$ is the end index of the predicted span.

\subsection{Fine-tuning and Inference}
In the fine-tuning phase, we regard the sum of the negative log likelihood of the true start and end indices as training loss,
\begin{equation}
    L = -\frac{1}{N}\sum_{l=1}^{N} \left[
    \log(p^{\textrm{s}}_{y_l^1}) + \log(p^{\textrm{e}}_{y_l^2})
    \right],
\end{equation}
where $N$, $y_l^1$, and $y_l^2$ denote the number of examples,
true start, and true end indices of the $l$-th example, respectively.
If answer type prediction is necessary, we add the cross entropy loss of the answer type to the training loss.
In the inference phase, the answer span $(s_{i}, e_{i})$
is calculated by dynamic programming, where the values of $p^{\textrm{s}}$ and $p^{\textrm{e}}$ are maximum and $1 \leq s_{i} \leq e_{i} \leq T$.

\begin{table*}[t]
    \centering
    \small
        \setlength{\tabcolsep}{4pt}
        \begin{tabular}{lccccc|cc|cc|c}
        \hline \hline
         & \multicolumn{5}{c|}{In-domain} & \multicolumn{2}{c|}{Out-of-domain} & In-domain & Out-of-domain & \\
         & Child. & Liter. & Mid-High. & News & Wiki & Reddit & Science & overall & overall &Overall \\
        \hline
        DrQA + PGNet & 64.2 & 63.7 & 67.1 & 68.3 & 71.4 & 57.8 & 63.1 & 67.0 & 60.4 & 65.1 \\
        BiDAF++ (3-ctx) & 66.5 & 65.7 & 70.2 & 71.6 & 72.6 & 60.8 & 67.1 & 69.4 & 63.8 & 67.8 \\
        FlowQA (1-ans) & 73.7 & 71.6 & 76.8 & 79.0 & 80.2 & 67.8 & 76.1 & 76.3 & 71.8 & 75.0 \\
        SDNet (single)  & 75.4 & 73.9 & 77.1 & 80.3 & \textbf{83.1} & 69.8 & 76.8 & 78.0	& 73.1 & 76.6 \\
        \hline
        BERT w/ 2-ctx & \textbf{76.0} & \textbf{77.0} & \textbf{80.5} & \textbf{82.1} & 83.0 & \textbf{72.5} & \textbf{79.6} & \textbf{79.8} & \textbf{75.9} & \textbf{78.7} \\
        \hline
        ConvBERT (single)  & - & - & - & - & - & - & - & 87.7 &	84.6 & 86.8 \\
        Google SQuAD 2.0 & \multirow{2}{*}{-} & \multirow{2}{*}{-} & \multirow{2}{*}{-} & \multirow{2}{*}{-} & \multirow{2}{*}{-} & \multirow{2}{*}{-} & \multirow{2}{*}{-} & \multirow{2}{*}{88.5} & \multirow{2}{*}{86.0} & \multirow{2}{*}{87.8} \\
        \ \ + MMFT (single) & & & & & & & & & & \\
        \hline \hline
    \end{tabular}
    \caption{The results on the CoQA test set of single models ($\textrm{F}_{1}$ score). Our BERT w/ 2-ctx model ranked 13th among all unpublished and published models (including ensemble) on the leaderboard at the submission time (April 13, 2019). The ConvBERT and the Google SQuAD 2.0 + MMFT are the current state-of-the-art models, but they are unpublished.}
    \label{tab:results_of_CoQA}
\end{table*}

\section{Experiment} \label{sec:experiment}
In this section, we evaluate our method on two conversational machine comprehension datasets,
QuAC \citep{QuAC} and CoQA \citep{CoQA}.
\subsection{Datasets and Evaluation Metrics}
Although CoQA is released as an abstractive CMC dataset,
\citet{QuAC_CoQA_comparison} shows that the extractive approach is also effective for CoQA.
Thus, we also use our extractive approach on CoQA.
To handle answer types in CoQA, 
we predict the probability distribution of the answer type (\textit{SPAN}, \textit{YES}, \textit{NO}, and \textit{UNANSWERABLE})
and replace the predicted span with "yes", "no", or "unknown" tokens
except for the "SPAN" answer type.
In QuAC, the unanswerable questions are handled as an answer span ($P$ contains a special token), and the type prediction for yes/no questions is not evaluated on the leaderboard. Therefore, we skip the answer type prediction step.

As evaluation metrics for CoQA, we use the $\textrm{F}_{1}$ score.
CoQA contains seven domains as paragraph contents:
children’s stories, literature, middle and high school English exams,
news articles, Wikipedia articles, science articles, and Reddit articles.
We report $\textrm{F}_{1}$ for each domain and the overall domains. 
On the other hand, as evaluation metrics of QuAC, we use not only $\textrm{F}_{1}$
but also the human equivalence score for questions (HEQ-Q) and for dialogues (HEQ-D)  \citep{QuAC}.
HEQ-Q represents the percentage of exceeding the model performance over the human evaluation for each question,
and HEQ-D represents the percentage of exceeding the model performance over the human evaluation for each dialogue.

\subsection{Comparison Systems}
We compare our model (BERT w/ k-ctx) with the baseline models and published models.
For QuAC, we use the reported scores of BiDAF++ w/ k-ctx \citep{QuAC} and FlowQA \citep{FlowQA}.
For CoQA, the comparison system is DrQA+PGNet \citep{CoQA}, BiDAF++ w/ x-ctx, FlowQA, and SDNet \citep{SDNet}.
Note that the scores of BiDAF++ w/ x-ctx on CoQA are reported by \citet{QuAC_CoQA_comparison}.
In addition, we use gold answers as the answer history,
except for the investigation of the effect of answer history.
More information on our implementation is available in Appendix \ref{sec:implementation_details}.

\subsection{Results}
\begin{table}[t]
    \centering
    \small
    \begin{tabular}{lccc}
        \hline \hline
        & $\textrm{F}_{1}$ & HEQ-Q & HEQ-D \\
        \hline
        BiDAF++ (2-ctx) & 60.1 & 54.8 & 4.0 \\
        FlowQA (2-ans)  & 64.1 & 59.6 & 5.8 \\ 
        \hline
        BERT w/ 2-ctx & \textbf{64.9} & \textbf{60.2} & \textbf{6.1} \\
        \hline
        ConvBERT (single) & 68.0 & 63.5 & 9.1 \\
        Bert-FlowDelta (single) & 67.8 & 63.6 & 12.1 \\
        \hline \hline
    \end{tabular}
    \caption{The results on the QuAC test set of single models. Our BERT w/ 2-ctx model ranked 1st among all unpublished and published models on the leaderboard at the submission time (March 7, 2019).  The ConvBERT and Bert-FlowDelta are the current state-of-the-art models, but they are unpublished.}
    \label{tab:results_of_quac}
\end{table}
\paragraph{Does our model outperform published models on both QuAC and CoQA?}
\tbl{results_of_CoQA} and \tbl{results_of_quac} show the results on CoQA and QuAC, respectively.
On CoQA, our model outperformed all of 
the published models regarding the overall $\textrm{F}_{1}$ score.
Although our model was comparable with SDNet for the Wikipedia domain, our model outperformed SDNet for the other domains.
On QuAC, our model also obtained the best score among the published models for all of the metrics
and obtained state-of-the-art scores on March 7th, 2019.

Our method uses the paragraph representations independently conditioned with each question and each answer. 
This model structure is suitable for the pre-trained BERT, which was trained with two input segments. Therefore, our model was able to capture the interaction between a dialogue history and a paragraph, and it achieved high accuracy.

\begin{table}[t]
    \centering
    \small
    \begin{tabular}{lccc}
        \hline \hline
        & \# contexts & CoQA & QuAC \\
        \hline
        BERT w/ 0-ctx & 0 & 72.8 & 55.0 \\
        \hline
        BERT w/ 1-ctx & 1 & 79.2 & 63.4 \\
        BERT w/ 2-ctx & 2 & 79.6 & \textbf{65.4} \\
        BERT w/ 3-ctx & 3 & 79.6 & 65.3 \\
        BERT w/ 4-ctx & 4 & 79.4 & 64.8 \\
        BERT w/ 5-ctx & 5 & \textbf{79.7} & 64.5 \\
        BERT w/ 6-ctx & 6 & 79.5 & 64.9 \\
        BERT w/ 7-ctx & 7 & \textbf{79.7} & 64.4 \\
        \hline \hline
    \end{tabular}
    \caption{The results with the number of previous contexts on the development set of QuAC and CoQA ($\textrm{F}_{1}$ score)}
    \label{tab:results_with_the_number_of_previous_context}
\end{table}
\paragraph{Does our model improve the performance when the number of previous contexts increases?}
\tbl{results_with_the_number_of_previous_context} shows the results with the number of previous contexts.
On both of the datasets, it was effective to use previous contexts.
However, on CoQA, the number of contexts had little effect on the score
even if the long context was considered.
On QuAC, the best score was obtained in the case of using two contexts, and the score decreased with more than two contexts.
As \citet{QuAC_CoQA_comparison} mentioned,
the topics in a dialogue shift more frequently on QuAC than on CoQA.
Thus, the previous context on QuAC can include the context that is unrelated to the current question, and this unrelated context can decrease the score.
This result suggests that it is important to select context that is related to the current question and not use the whole context in any cases.

\begin{table}[t]
    \centering
    \small
    \begin{tabular}{lcc}
        \hline \hline
         & CoQA & QuAC \\
         \hline
        BERT w/ 0-ctx & 72.8 & 55.0 \\
        BERT w/ 2-ctx (gold ans.) & \textbf{79.6} & \textbf{65.4} \\
        \ \ w/o question history & 78.0 & 64.7 \\
        \ \ w/o answer history & 77.7 & 59.3 \\
        BERT w/ 2-ctx (predicted ans.) & 77.2 & 56.7 \\
        \hline \hline
    \end{tabular}
    \caption{Ablation study on the development set of QuAC and CoQA ($\textrm{F}_{1}$ score)}
    \label{tab:ablation_study}
\end{table}
\begin{figure}[t]
    \centering
    \includegraphics[width=8cm]{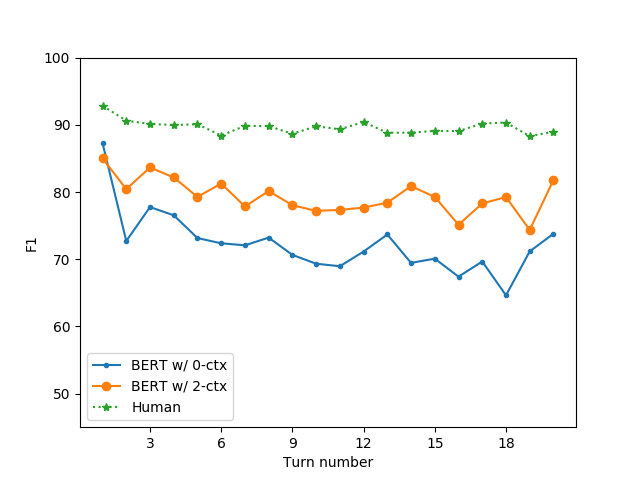}
    \caption{The $\textrm{F}_1$ scores with turn number on CoQA development set}
    \label{fig:CoQA_F1_turn_number}
\end{figure}

\paragraph{Which is more important, the question history or the answer history?}
\tbl{ablation_study} shows the contribution of the dialogue history.
We can see from the results that the model performance decreased significantly when we removed the gold answer history on QuAC. 
In dataset collection, CoQA allows the asker to see the evidence paragraph. On the other hand, the asker in QuAC cannot see the evidence paragraph. As a result, questions in QuAC are far from the phrases in the passage and are less effective in improving the model performance. For CoQA, the model could substitute the question history for the gold answer history. The model performance did not decrease significantly when we remove the answer history.


\paragraph{Does our model maintain the performance when using the predicted answer history?}
In actual conversation, the gold answer history may not be given in the CMC model. In this experiment, we trained the models with the gold answer history and evaluated the model with the predicted answer history.

As shown in \tbl{ablation_study}, 
when using the predicted answer 
history, the model performance 
decreased significantly on QuAC.
This result also suggests that the model can substitute the question history for the gold answer history in CoQA.
We think 
the CMC setting where the history of questions posed by an asker that does not see the evidence paragraph is given and the gold answer is not given for input is a more realistic and important setting.


\begin{figure}[t]
    \centering
    \includegraphics[width=8cm]{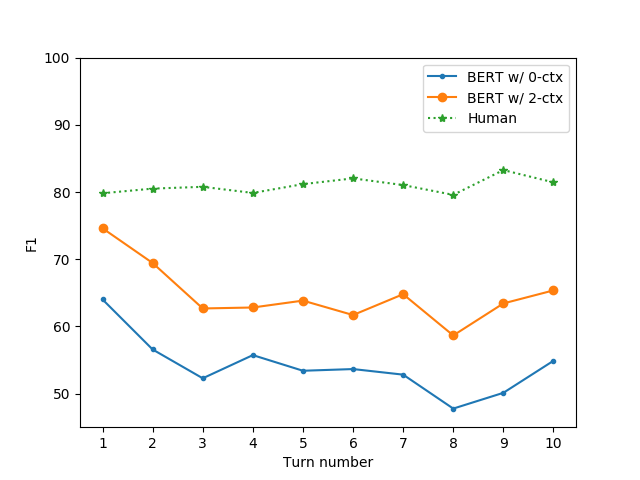}
    \caption{The $\textrm{F}_1$ scores with turn number on QuAC development set}
    \label{fig:QuAC_F1_turn_number}
\end{figure}
\paragraph{Does our model performance approach human performance as the dialogue progresses?}
We calculated $\textrm{F}_1$ scores over the turns, where the data in each turn contained more than 100 question/answer pairs.
\fig{CoQA_F1_turn_number} and \fig{QuAC_F1_turn_number} show that
the score was lower than human performance over all turns on both datasets
and that the score with context was higher than that without context on both datasets, except for the first question on CoQA.
This result indicates that there is still room for improvement with long turn questions.


\section{Related Work} \label{sec:related_work}
QuAC \citep{QuAC} and CoQA \citep{CoQA} were released as the CMC dataset.
On QuAC, the answers are extracted from source paragraph as spans.
On CoQA, the answers are free texts based on span texts extracted from the source paragraph.
On these datasets, the baseline models were 
based on conventional models for single-turn machine comprehension
such as BiDAF \citep{BiDAF} and DrQA \citep{DrQA}.
For QuAC, \citet{QuAC} extended BiDAF (an extractive machine comprehension model) to BiDAF++ w/ x-ctx
by concatenating word embeddings of the source paragraph and embeddings of previous answer span indexes.
For CoQA, \citet{CoQA} proposed DrQA+PGNet as an abstractive method
by concatenating previous questions and previous answers with special tokens.
However, most of the recently published methods about CoQA were extractive approaches,
since the abstractive answers on CoQA are based on span texts in the paragraph and \citet{QuAC_CoQA_comparison} shows that the extractive approach is also effective for CoQA.
\citet{FlowQA} proposed FlowQA for both QuAC and CoQA 
by stacking bidirectional recurrent neural networks (RNNs) over the words of the source paragraph
and unidirectional RNNs over the conversational turns.
\citet{SDNet} proposed SDNet for CoQA
by regarding the concatenation of previous questions and answers as one query.

Most recently, BERT \citep{BERT} was proposed as a contextualized language representation
that is pre-trained on huge unlabeled datasets.
By fine-tuning a supervised dataset, BERT obtained state-of-the-art scores
on various tasks including single-turn machine reading comprehension datasets such as SQuAD \citep{SQuAD_v1}.
Since the relationship between words 
can be captured in advance,
pre-training approaches such as BERT and GPT-2 \citep{GPT2} can be useful especially for tasks
with a small amount of supervised data.
For QuAC and CoQA, many approaches on the leaderboard\footnote{\burl{https://quac.ai/}}$^{,}$\footnote{\burl{https://stanfordnlp.github.io/coqa/}}
use BERT, including SDNet.
However, SDNet uses BERT as contextualized word embedding without updating the BERT parameters.
This is one of the differences between SDNet and our model.

\section{Conclusion} \label{sec:conclusion}
In this paper, we propose a simple but effective method based on a fine-tuning approach with BERT
for a conversational machine comprehension (CMC) task.
Our method uses questions and answers simply as the input of BERT to model the interaction between the paragraph and each dialogue history independently and outperformed published models on both QuAC and CoQA.

From detailed analysis, we found that the gold answer history, which may not be given in real conversational situations, contributed to the model performance most on both datasets.
We also found that the model performance on QuAC decreased significantly when we used predicted answers instead of gold answers. On the other hand, we can substitute the question history for the gold answer history on CoQA.
For future work, we will investigate a more realistic and more difficult CMC setting, where the history of questions posed by the asker that does not see the evidence paragraph is given and the gold answer is not given for input. We will also investigate how to obtain related and effective context for the current question in the previous question and answer history. 




\bibliography{references}
\bibliographystyle{acl_natbib}

\appendix

\section{Implementation Details} \label{sec:implementation_details}
We used the BERT-base-uncased model implemented by PyTorch 
\footnote{\burl{https://github.com/huggingface/pytorch-pretrained-BERT}}.
We used a maximum sequence length of 384, document stride of 128,
maximum query length of 64, and maximum answer length of 30.
The optimizer was Adam \citep{Adam} with a learning rate of 3e-5, $\beta_{1} = 0.9$, $\beta_{2} = 0.999$,
L2 weight decay of 0.01, learning rate warmup over the first 10 \% of training steps,
and linear decay of the learning rate.
The number of training epochs was 2.
The batch size of training was 8 or 12.
In the case of QuAC, we used dialogs whose paragraphs have under 5,000 characters.
In the case of CoQA, we followed \citet{FlowQA} and regarded a span with maximum $\textrm{F}_{1}$ overlap
with respect to given abstractive answers as gold answers during training.
We used four NVIDIA Tesla V100 32GB GPUs.

\end{document}